\pdfoutput=1

\documentclass[letterpaper, 10 pt, conference]{ieeeconf}  

\usepackage{times}
\usepackage{epsfig}
\usepackage{graphicx}
\usepackage{amsmath}
\usepackage{amssymb}
\usepackage{subfigure} 
\usepackage{hhline}

\usepackage{bbding}
\usepackage{pifont}
\usepackage{wasysym}
\usepackage{amssymb}
\usepackage{verbatim}

\usepackage{chngcntr}
\counterwithin*{paragraph}{subsection}

\IEEEoverridecommandlockouts                              

\renewcommand{\epsilon}{\varepsilon}

\def\utilde#1{\mathord{\vtop{\ialign{##\crcr
$\hfil\displaystyle{#1}\hfil$\crcr\noalign{\kern1.5pt\nointerlineskip}
$\hfil\tilde{}\hfil$\crcr\noalign{\kern1.5pt}}}}}

\newcommand{\hide}[1]{}

\newcommand{\Gellert}[1]{{\color[rgb]{0,0,0}{#1}}}

\newcommand{\bP}{\mathbf{P}}

\newcommand{\camready}[1]{{\color[rgb]{0,0,0}{#1}}}

\usepackage[pagebackref=true,breaklinks=true,letterpaper=true,colorlinks,bookmarks=false]{hyperref}

\title{Deep Multi-Sensor Lane Detection}

\author{Min Bai$^{*,1,2}$ \qquad Gellert Mattyus$^{*,1}$ \qquad Namdar  Homayounfar$^{1,2}$ \qquad Shenlong Wang$^{1,2}$ \\ Shrinidhi  Kowshika Lakshmikanth$^{1}$ \qquad Raquel Urtasun$^{1,2}$
\thanks{$^1$ MB, GM, NH, SW, SKL, and RU is with Uber Advanced Technologies Group,{ \tt\small \{mbai3, gmattyus, namdar, slwang, klshrinidhi, urtasun\}@uber.com}}
\thanks{$^2$ MB, NH, SW, RU is with University of Toronto, {\tt\small \{mbai, namdar, slwang, urtasun\}@cs.toronto.edu}}
}

\begin{document}


\maketitle

\begin{abstract}
Reliable and accurate lane detection has been a long-standing problem in the field of  autonomous driving. In  recent years, many approaches have been developed that use images (or videos) as input and reason in image space. In this paper we argue that accurate image estimates do not translate to precise 3D lane boundaries, which are  the input required by modern motion planning algorithms. To address this issue, we propose a novel  deep neural network  that  takes advantage of both LiDAR and camera sensors and  produces very accurate estimates directly in 3D space. We demonstrate the performance of our approach on both highways and in cities,  and show very accurate estimates in complex scenarios such as heavy traffic (which produces occlusion), fork, merges and intersections. 
\end{abstract}


\section{Introduction}

Lane detection is one of the fundamental problems in autonomous driving. To drive safely, the vehicle must reliably detect the boundaries of its current lane for accurate localization, while also estimating the nearby lanes for maneuvering. Accurate detections of lane regions can help reduce ambiguities when detecting and tracking other traffic participants. Furthermore, it is an essential component for the automatic creation of high definition maps, as well as error checking and change detection in existing maps. 

\begin{figure*}[t]
\begin{center}

\begin{subfigure} \centering \includegraphics[height=3.0cm]{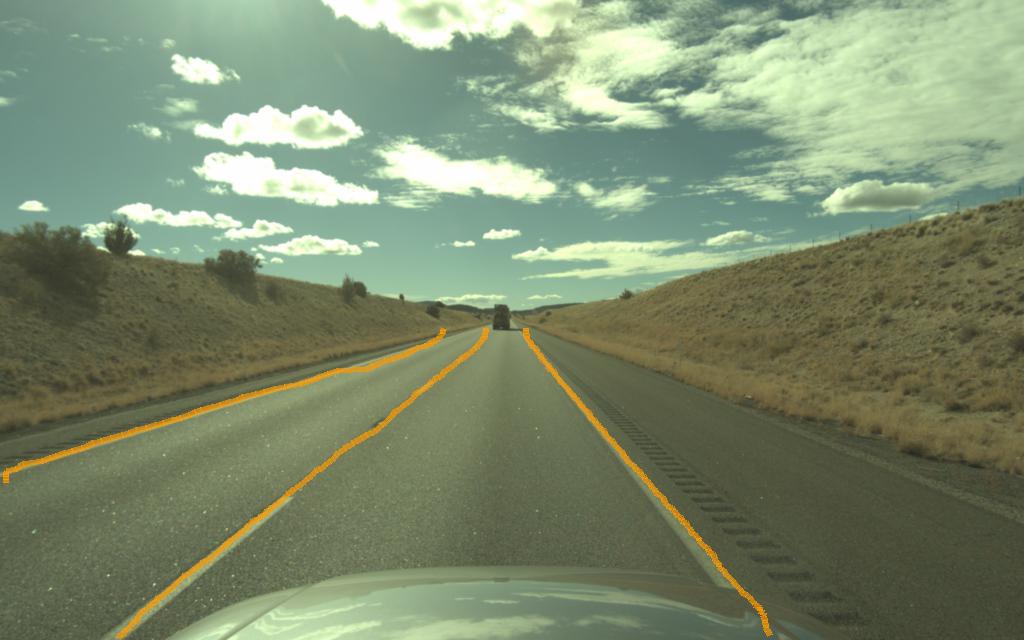} \end{subfigure}
\begin{subfigure} \centering \includegraphics[height=3.0cm]{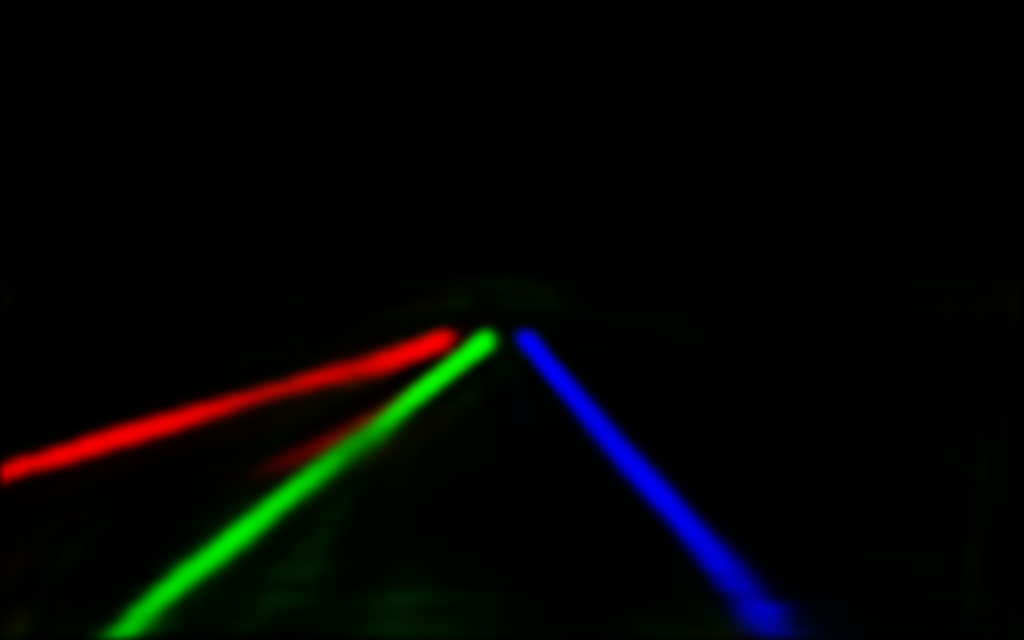} \end{subfigure}
\begin{subfigure} \centering \includegraphics[height=3.0cm]{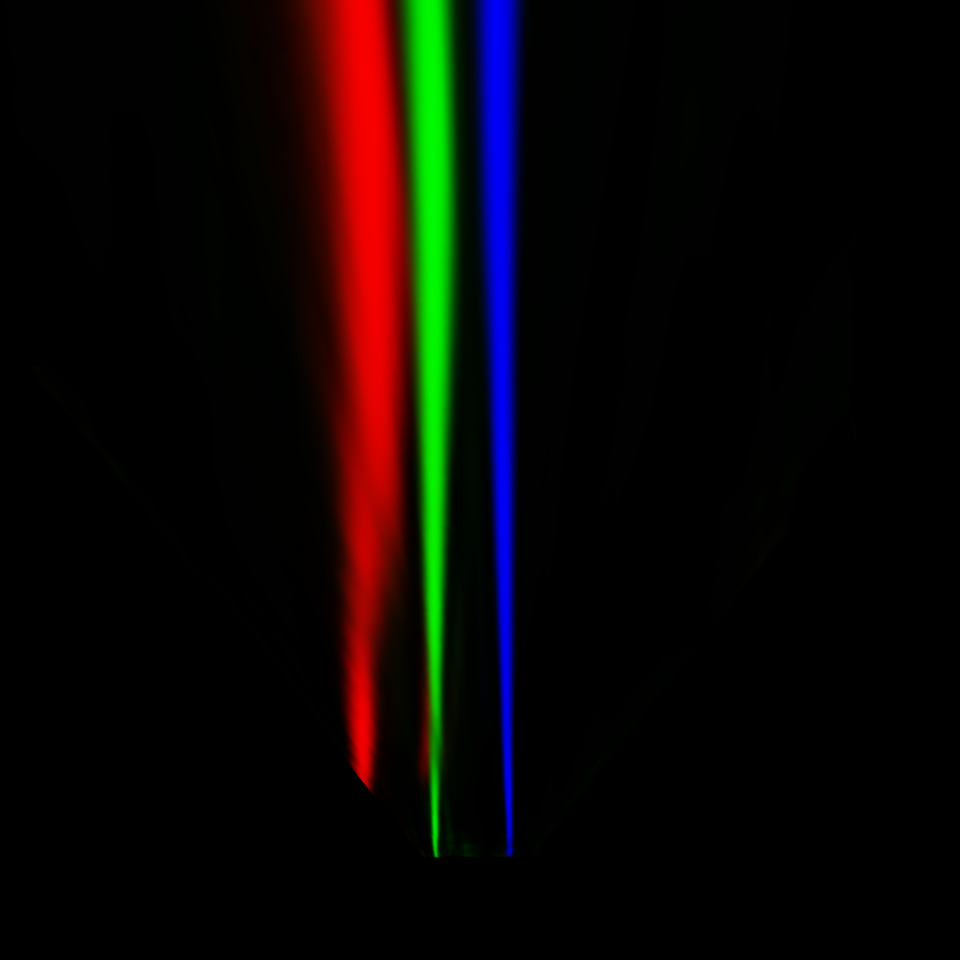} \end{subfigure}
\begin{subfigure} \centering \includegraphics[height=3.0cm]{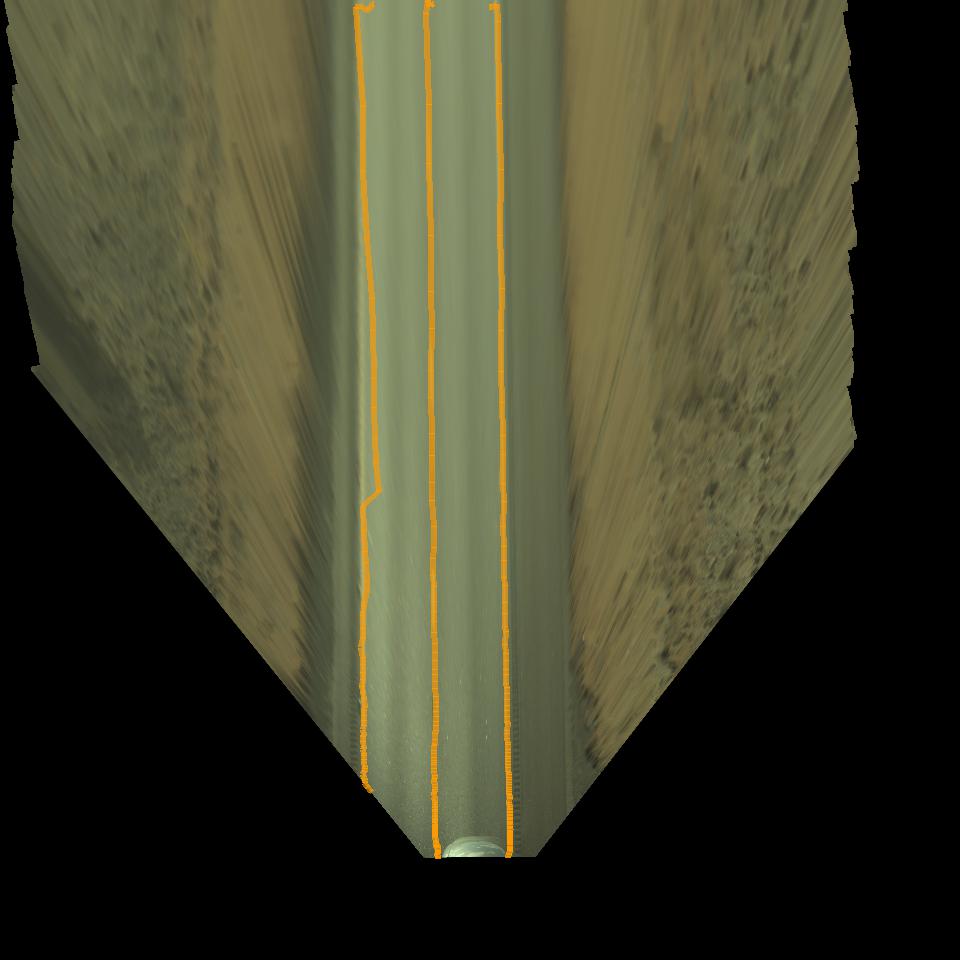} \end{subfigure}
\caption{Illustration of the necessity of reasoning about lane detections in the overhead view. The input camera image overlaid with the predictions from \cite{sensetime} is shown on the left. The lane probability output of the model in camera view is shown next, followed by the re-projection of the probability output into overhead view. The lane detections re-projected into overhead view is on the right. Although the lane detections in camera view appear quite accurate, the shortcomings are clearly visible in the overhead view, as the probability output becomes diffuse with increasing distance, and the detected lane boundaries become less accurate. }\label{fig:reproj_to_BEV}
\end{center}
\end{figure*}

Most approaches to lane detection take an image (or a video) as input and estimate the lane boundaries in image space \cite{haloi2015, sensetime, kim2017,huval2015,bailo2017, hur2013}.  
In this paper, we argue that accurate results in the camera view may not imply accurate estimates in 3D space.  This is due to the fact that the perspective projection of the camera causes the spatial resolution in 3D to decrease drastically with distance from the camera. 
This is a big issue for modern self driving vehicles as motion planners require a birds eye view representation of the lane topology. 
We refer the reader to Fig.~\ref{fig:reproj_to_BEV} for an illustration of this problem. The figure shows sample outputs of the state-of-the-art camera based lane detection approach of \cite{sensetime}. 
Despite very accurate results in  camera perspective, the results in 3D are not very accurate. 


Several approaches that directly use 3D sensors such as LiDAR have been proposed \cite{kammel2008lidar, lindner2009,thuy2010,hata2014, huangprobabilistic, huang2009finding}. Although LiDAR gives an unambiguous measurement of 3D points, it is spatially much more sparse when compare to images, especially at long-range. Consequently, many of the proposed methods relied on handcrafted features and strong assumptions, \textit{e.g.~}fitting parabolic curves. Additionally, they often fall short in reliability, especially in less ideal scenarios like heavy occlusion. While camera-based techniques have been fairly extensively explored, the possibility to exploit the rapid progress in deep learning to boost the accuracy and reliability of lane detectors using LiDAR or multi-sensor input remains open.

In this paper we propose a novel  deep neural network  that  takes advantage of both LiDAR and camera sensors and  produces very accurate estimates directly in 3D space. 
Our network can be learned end-to-end, and can produce reliable estimates with any combination of sensors. 
Importantly, our approach is very efficient and runs in as little as 70ms on a single Titan Xp GPU when using all sensors. 
We demonstrate the effectiveness of our approach on two large scale real world datasets containing both highway and city scenes, and show significant improvements over the state-of-the-art. 

\begin{figure*}[t]
\begin{center}

\begin{subfigure} \centering \includegraphics[width=\textwidth]{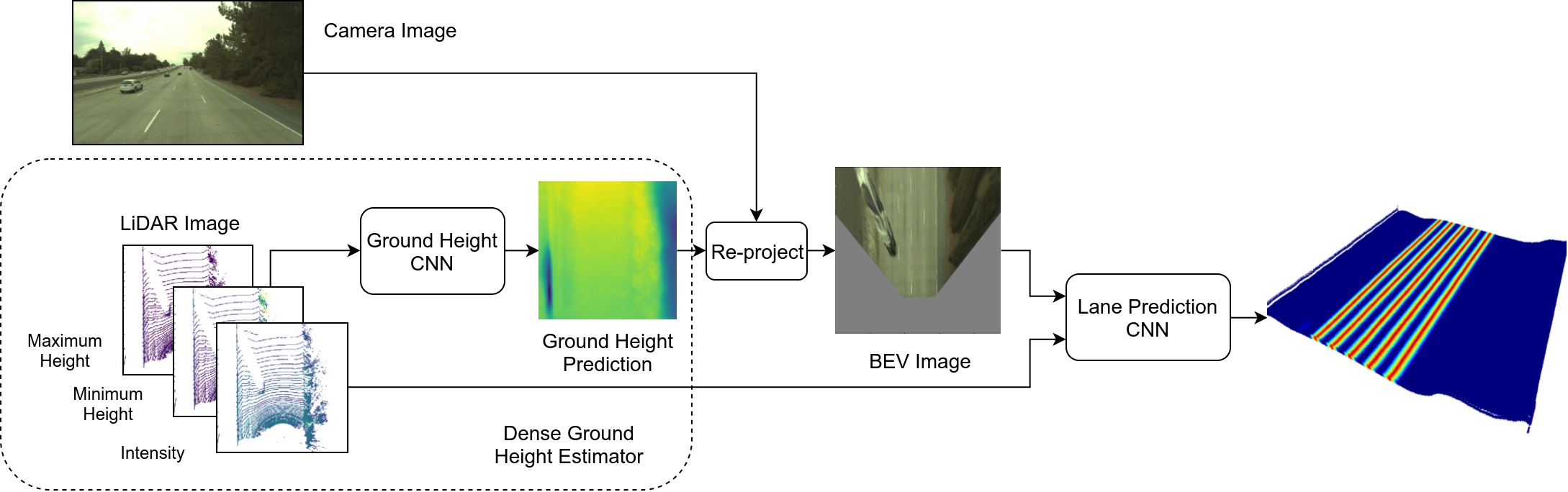} \end{subfigure}

\caption{Complete network architecture. Our model takes as input information extracted from LiDAR sweeps, and predicts a dense ground height. The input RGB camera image is projected onto the dense ground, and is combined with the LiDAR information to produce lane detections in overhead view 3D view. }\label{fig:network}
\end{center}
\end{figure*}

\section{Related Work}
This section discusses the various techniques that have been proposed to tackle the lane detection task. The techniques are grouped by their themes. 

\paragraph{Camera view lane detection}

A commonly seen family of solutions reason in the first person perspective of a vehicle. This includes a number of traditional computer vision techniques based on feature extraction using hand-crafted methods to identify likely locations of lane markings. Such methods include the use of various edge detection filters \cite{haloi2015}, corner detection features \cite{wu2012}, Haar-like features \cite{jung2013}, and clustering \cite{zhang2013}. This is then often followed by line or spline fitting with techniques such as RANSAC to yield final lane detections. Some techniques further use conditional random fields to refine the outputs \cite{hur2013}. However, these techniques are often held back by their inability to reason in abstract domains in the presence of heavy occlusion by other traffic participants, varying lighting conditions between day and night scenes, and weather conditions. Additionally, due to the perspective transformation by the camera, far away road and lane markings are simultaneously heavily distorted and reduced in resolution. 

Recent advances in convolutional neural networks (CNN) have led to a drastic increase in performance in various 2D computer vision tasks. Many methods have been proposed that leverage these powerful models for feature extraction or direct lane detection \cite{sensetime, kim2017,huval2015,bailo2017}.

Some techniques have been proposed which attempt to detect lanes in the camera view, but additionally use regularities in the 3D space to guide the detections. For example, \cite{lee2017} exploits the fact that lanes are mostly parallel lines in 3D and converge at a vanishing point in the projected image. They use the 2D detections of these vanishing points to further guide their CNN-based lane detection model. Additionally, \cite{gurghian2016} explored an interesting framework which uses transfer learning methods to improve lane detection performance on different datasets. 

\paragraph{Overhead view lane detection}

As will be shown later in our work, reasoning about lanes in the overhead view domain has significant advantages. This setting has previously been explored by \cite{he2016}, where the authors use a siamese CNN to simultaneously reason about both overhead and camera view input. An earlier work by \cite{kheyrollahi2012} uses perspective mapping to project camera images into an assumed fixed ground plane and a relatively simple neural network model for the related task of road symbol detection. There, they demonstrated the benefits of reasoning in the overhead view as it reduces the perspective distortion of far away road symbols. This scheme is followed by other authors, including \cite{hyeon2016}. However, these schemes neglect the natural swaying of the ego-vehicle due to its suspension system, as well as inherent sloping of the road. This limits the spatial accuracy of their detections. In contrast, we propose an end-to-end trained deep learning model that predicts an accurate ground height estimation as an intermediate step onto which the camera image is projected. 

\paragraph{LiDAR-based lane detection}

Several techniques have been proposed using LiDAR measurements as the input \cite{kammel2008lidar, lindner2009,thuy2010,hata2014}. However, these techniques largely do not leverage the recent advances in deep learning. Consequently, they either assign each point a label resulting in only sparse labels, or using some underlying assumption, such as parabolic curve to fit a lane model. Other works have successfully applied CNNs to the task of road segmentation, especially in the overhead view \cite{caltagirone2017}. Our work takes this one step further to show that information from LiDAR is sufficient for dense and accurate lane boundary detections as well. 

\paragraph{Multi-sensor lane detection}
Previous works also exploit the use of multiple sensors to boost the performance of lane detection and tracking \cite{huangprobabilistic, huang2009finding}. However, lane perception only uses the cameras, while LiDAR is used for supplementary tasks such as obstacle masking and curb fitting. Unlike these methods, our approach aggregates information from both sensors to detect lanes in the same output space.

\section{Multi-model End-to-End Lane Detection}

In this section, we propose a new model that produces accurate and reliable lane boundaries  in 3D space. Our model combines the strength of camera's dense observations as well as LiDAR's unambiguous 3D measurements. In the following, we first define the input and output parameterization, followed by a description of the architecture of our proposed end-to-end learnable model. We then discuss each individual component in detail as well as   how to train our model.

\subsection{Parameterization of the Input and Output}


Our approach is inspired by the observation that performing lane detection in image space can result in large inaccuracies in 3D (especially at long range).  To address this issue, we phrase the problem  as a dense pixel-wise prediction task in 3D vehicle coordinates, regardless of the type of sensor input employed. 
This enables the lane detection to directly deliver lane boundaries in the same space in which the motion planning and control system operate. 

Most existing pixel-wise CNN-based approaches (e.g. \cite{sensetime}) attempt to predict whether or not a given pixel belongs to a lane boundary. 
However, they disregard the distinction between cases where the detected lane boundary is slightly shifted compared to the ground truth, and cases where the detection is completely spurious or missing. To address this problem, we phrase the output space as the minimum Euclidean distance to the nearest lane boundary at each location. This allows the model to explicitly reason about the relationship of each 3D location and lane boundaries, providing an output that encapsulates richer information than class labels. Moreover, the loss is more forgiving to small offsets between the prediction and ground truth. To aid the network's learning, we make a small modification to the distance transform. In particular, we eliminate the unbounded regression problem by thresholding the target value, and invert the distance transform by subtracting it from the threshold itself. As a result, the training target is equal to the threshold at the lane boundary, and decays outward to zero.

Our model takes two inputs: a LiDAR point cloud  as well as an image of the scene.  In particular, we bring the point clouds from five consecutive frames to a common reference frame by correcting for ego-motion.  We then rasterize the  point clouds to a 3-channel image in Bird's Eye View (BEV) encoding both  the intensity and height of the highest point as well as the height of the lowest point in each discretization bin. \camready{It is important to note that there may be independently moving objects such as other vehicles in the scene whose motion would not be compensated for. However, by encoding the highest and lowest height in each discretization bin, the network is informed of the regions that contain moving objects. Using this, the network can learn to ignore distracting data, and interpolate where necessary. } The second input is the RGB camera image. 
We refer the reader to Fig. \ref{fig:network} for an illustration of our model. In the next section, we describe our multi-sensor neural network in details.

\subsection{Network Architecture.} 


To achieve the goal of simultaneously leverage information from the first person view camera image and the BEV LiDAR image, we must align their domain by re-projecting the camera image into BEV as well. 
Since our camera is calibrated and fixed to the vehicle, the projection of a point $\bP_v= (x_v, y_v, z_v)^T$  in the vehicle frame is defined by the projection matrix $C=K [R_v | t_v]$, where $K$ is the camera calibration matrix and $R_v, t_v$ is the rotation and translation from vehicle to the camera coordinate system. 
Because the vehicle frame is fixed (i.e. $x_v, y_v$ are constants), only the $z_v$ elevation of the ground has to be estimated online to define the projection from ground into the camera. 

The simplest solution is to assume that the road is flat and parallel to the vehicle frame, in which case $z_v$ would be a constant as well. 
Unfortunately, real world roads often have significant slopes over large distances. Moreover, the elevation and pitch of the vehicle relative to the ground is constantly changing due to the vehicle's suspensions. \camready{Because of the oblique angle, ground regions far away from the vehicle are covered by very few LiDAR measurements. As well, many LiDAR readings do not come from the ground, but rather various obstacles and other traffic participants. From experimentation, we find that these pose significant difficulties when attempting to use traditional robust estimators such as RANSAC to produce a dense ground height estimate. }
Instead, in this paper we take advantage of deep learning and design a network that estimates dense ground from the sparse LiDAR point cloud, \camready{which learns to simultaneously ignore objects above the ground, as well as extrapolate to produce a dense ground surface. }

\noindent\textbf{Ground Height Estimator}: We  use a fast convolutional neural net (CNN) based on ResNet50 \cite{he2016_resnet} to predict a dense ground height from the LiDAR input. We reduce the feature dimensions of ResNet50 by a factor of 2 for scales 1 and 2, and a factor of 4 for scales 3 to 5. Additionally, we remove one convolutional block from each of scale 2, 3, and 5. This is followed by large receptive field average pooling inspired by \cite{zhao2017}. In particular, we produce three additional feature volumes using pooling with receptive field and stride sizes of 10, 25, and 60, before concatenating the feature volumes with the original output of scale 5. This gives the network the capability to propagate information about the ground height from regions with high LiDAR point density to regions with low density. Next, the feature volume undergoes 3 additional bottleneck residual blocks, before being upsampled using interleaved transposed convolutions and basic residual blocks. The result is a dense ground height image in BEV of size $960 \times 960$. This corresponds to a region of $48 \times 48$m at 5cm per pixel. 

\noindent\textbf{Camera Image Re-projection}: We project the camera image to the estimated ground surface using a differentiable warping function \cite{jaderberg2015}. This produces an image warped onto the dense ground height prediction described above. Note that the re-projection process does not explicitly handle 3D occlusions in the ground plane, with pixels of objects above the ground being projected onto the ground surface. However, as markings guiding lane detections are on the ground plane, the locations of detected lanes will be largely unaffected. The result is a re-projected camera image with the same size as the predicted dense ground image above. 

\noindent\textbf{Combining Re-projected Camera Image with LiDAR}: we use a second CNN to leverage both the re-projected camera image and the LiDAR input to produce pixel-wise lane detection results. \camready{This module takes as input LiDAR data in the same format as the ground height estimator.} We design a second CNN that is largely identical to that used in the dense ground height estimation, with the exception that the feature dimension of scales 3 to 5 are only reduced by a factor of 2 relative to the original ResNet50, with no blocks removed. Moreover, we duplicate scales 1 to 3 at the input without weight sharing such that the re-projected camera image and the LiDAR images are passed into separate input branches. The two streams of information are concatenated at scale 4. The output of this network is our distance transform estimates.

\subsection{Model Learning} 

The parameters $\Theta$ of the overall model are optimized by minimizing a combination of the lane detection loss and the ground height estimation loss: 

\[
l_\text{model}(\Theta) = l_\text{lane}(\Theta) + \lambda l_\text{gnd}(\Theta)
\]

The lane detection loss $l_\text{lane}$ is defined by 
\[
l_\text{lane}(\Theta) = \sum_{p \in \text{Output Image}} \|(\tau - \text{min} \{d_\text{p,gt}, \tau\}) - d_\text{p,pred}\|^2 
\]
where $d_\text{p,gt}$ and $d_\text{p,pred}$ are the ground truth and predicted distance transform values for pixel $p$, respectively. $\tau$ is a threshold used to cap the distance transform values to a range between $[0, \tau]$, as it is unnecessary for the CNN to produce exact distance transform values for regions far away from any lane boundary. Additionally, this is inverted such that the maximum predicted value $\tau$ occurs at the lane boundary, and linearly decreases outward to $0$. This removes the need for the network to predict a sharp drop-off in distance transform values at the thresholding boundary from $\tau$ to $0$. 

The ground height estimation loss $l_\text{gnd}$ is defined by 

\[
l_\text{gnd}(\Theta) = \sum_{p \in \text{Output Image}} \|z_\text{p,gt} - z_\text{p,pred}\|
\]
In this case, we select the L1 loss to encourage smoothness in the output.

\section{Experiments}

This section discusses the details of our dataset, evaluation metrics and results. 

\subsection{Datasets}

We evaluate our approach on two datasets collected in North America. The first dataset includes highway traffic scenes with 22073, 2572, and 5240 examples in training, validation, and test set, respectively. The second dataset includes city scenes from a medium sized city, with 16918, 2607, and 5758 examples in the training, validation, and test sets, respectively. The annotations includes the ground truth lane boundaries in both the camera view and overhead view. We use HD maps \camready{which contain dense 3D measurements and annotated lane graphs together with a localization system} to  provide us with dense ground height ground truth for each example. We plan to release this dataset.


\begin{figure}[]
\begin{center}

\begin{subfigure} \centering \includegraphics[width=\linewidth]{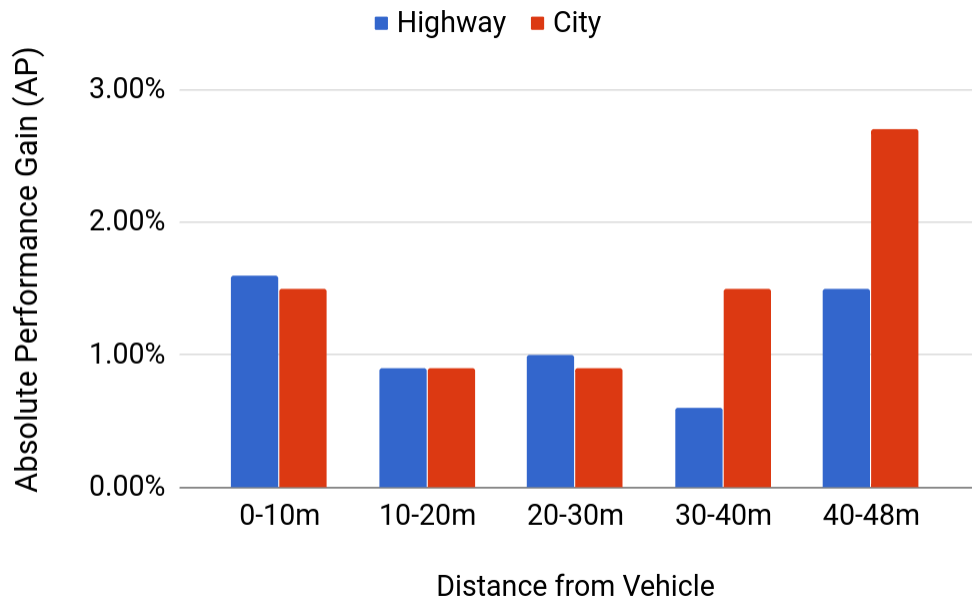} \end{subfigure}
\caption{Difference between the AP at various distances from vehicle of our LiDAR+Camera model versus the LiDAR-only model. It is evident that the LiDAR+Camera model outperforms the LiDAR-only model at all distances, with the advantage further increasing with distance. }\label{fig:distance}
\end{center}
\end{figure}


\newcommand{\resfigheight}{2.3}
\newcommand{\putnewimagerow}[2]{
\begin{subfigure} \centering \includegraphics[height=\resfigheight cm]{figures/results/#2/#1_f_proj_camera_img.jpg} \end{subfigure}
\begin{subfigure} \centering \includegraphics[height=\resfigheight cm]{figures/results/#2/#1_a_lidar_img.jpg} \end{subfigure}
\begin{subfigure} \centering \includegraphics[height=\resfigheight cm]{figures/results/#2/#1_c_lane_bev_gt_image.jpg} \end{subfigure}
\begin{subfigure} \centering \includegraphics[height=\resfigheight cm]{figures/results/#2/#1_b_out_lanes_ortho.jpg} \end{subfigure}
\begin{subfigure} \centering \includegraphics[height=\resfigheight cm]{figures/results/#2/#1_d_warped_camera_img.jpg} \end{subfigure}
}

\newcommand{\imagecaption}{
From left to right: (1) The detected lane distance transforms are naively thresholded are projected into the camera image using the predicted ground height. Lanes are predicted up to 48m ahead. (2) The LiDAR image. (3) The ground truth lanes. (4) The distance transform output of the model. 
(5) The camera warped to top-down view using the predicted ground overlaid with the predicted lanes (red).
}

\begin{figure*}[t]
\begin{center}

\putnewimagerow{18c88acb-275b-4423-f888-dfb8c20290401187655669777451}{cam_ready} \\
\putnewimagerow{273de478-15a3-46a3-f3c5-320cb18d30561185161775015828}{cam_ready} \\

\putnewimagerow{47aae0cf-186e-4837-f2c1-b8422b74bc771183764393085209}{cam_ready} \\
\putnewimagerow{ff341476-03d2-4d71-f91c-c2bc46e8771a1181404264352164}{cam_ready} \\
\putnewimagerow{4016d818-554a-4cfc-d71b-3261c401328d1200418444983691}{cam_ready} \\
\putnewimagerow{fd1deb39-083c-4265-f543-c58372dab5c71194900153411162}{cam_ready}

\caption{Output of our method using both LiDAR and camera on highways.
\imagecaption \Gellert{Lane detection can be highly non-trivial: vehicles can occlude the lane markings (rows 1-3), there may be distracting lines (row 5), and rain on both the ground and the camera sensor (last row). However, our model is able to leverage to multiple sources of information to produce high quality lane detections.}}\label{fig:results_hwy}
\end{center}
\end{figure*}

\begin{figure*}[t]
\begin{center}

\putnewimagerow{0a00f429-a971-d760-e869-77fa2befbff91190669918000001}{126_PHX_GroundPredictionCameraImageLidar2Branch_025000} \\

\putnewimagerow{0a4a6ffb-c2e2-3062-8c59-4d6d0c6dbbb11190061438333336}{126_PHX_GroundPredictionCameraImageLidar2Branch_025000} \\
\putnewimagerow{0e0a67c3-338d-4be9-64c7-24bff51a0ed71190137507000001}{126_PHX_GroundPredictionCameraImageLidar2Branch_025000} \\
\putnewimagerow{0a4a6ffb-c2e2-3062-8c59-4d6d0c6dbbb11190063207733331}{126_PHX_GroundPredictionCameraImageLidar2Branch_025000} \\

\putnewimagerow{0a4a6ffb-c2e2-3062-8c59-4d6d0c6dbbb11190064657133332}{126_PHX_GroundPredictionCameraImageLidar2Branch_025000} \\
\putnewimagerow{1b920920-23b4-37b2-057c-d4fde60118fd1190390903600002}{126_PHX_GroundPredictionCameraImageLidar2Branch_025000}


\caption{Output of our method using both LiDAR and camera in urban areas.
\imagecaption}\label{fig:results_phx}
\end{center}
\end{figure*}

\begin{figure*}[t]
\begin{center}

\putnewimagerow{fd1deb39-083c-4265-f543-c58372dab5c71194902381150041}{cam_ready/failures} \\
\putnewimagerow{72b1b868-bed7-498e-eee4-62165fe8862a1201992661243761}{cam_ready/failures} \\


\putnewimagerow{1b920920-23b4-37b2-057c-d4fde60118fd1190392592333332}{126_PHX_GroundPredictionCameraImageLidar2Branch_025000}


\caption{Failure cases due to \Gellert{rain}, occlusions and missing lane markings.
\imagecaption}\label{fig:results_failure}
\end{center}
\end{figure*}

\subsection{Experimental Setup}

Here, we describe the details of our training process. We set the loss mixing factor $\lambda = 20$. All models are trained using the ADAM \cite{kingma2015} optimizer with a learning rate of $1e-4$ and a weight decay of $1e-4$ with a batch size of $8$ until convergence as determined by the validation set loss. 
Additionally, we select the distance transform value threshold of $\tau=30$ for all models on highway scenes, and reduced it to $\tau=20$ for all models in city scenes to account for the closer spacing of lane boundaries in the latter scenario. This corresponds to $1.5$m and $1.0$m in the overhead view, respectively. 
\Gellert{We further improve the variations in our dataset by augmenting our dataset. We randomly adjust brightness, contrast, saturation and hue of the camera image, and randomly rotate the birds eye view images. To avoid artifacts in the LiDAR image (e.g. interpolate the height between empty and occupied pixels) we first rotate the 3D points prior to rasterization.}

\subsection{Metrics} 

We use two sets of metrics to measure the performance of our model. The first set of metrics directly compares the predicted and ground truth inverted truncated distance transforms over the output prediction by computing the L1 and L2 distances. These metrics naturally penalize both false positive and false negative detections of the lanes at a (overhead) pixel level. 

Additionally, we use a simple method to extract a line-based lane boundary representation by thresholding and binarizing the predictions at a fixed value. The thresholds are selected to be $20$ for the highway model, and $15$ for the city model, which are 10 px and 5 px away from the lane boundary, respectively. The result is skeletonized via binary erosion, which is then compared to the ground truth lane boundaries by computing the average precision scores at pixel thresholds ranging from 1 to 9. In the overhead image view, this corresponds to physical distances of 5 cm to 45 cm. Moreover, we evaluate the exact precision and recall values at 5 pixel deviation, or 25 cm in the overhead view. This is a stringent requirement, as in comparison the typical lane marking itself is only 15 cm wide. 

Finally, we evaluate the correctness of the detected lane topology by finding the number of connected components in the skeletonized prediction, and comparing it with the number of lanes present in the ground truth by computing the absolute difference. This statistic is averaged across all test examples. 
All results are reported in the test set. 

\subsection{Analysis}

We present qualitative and quantitative results of our method in this section. The performance of our complete model in highway and city scenes can be found in the third and sixth rows of Table \ref{sensor_comparison}. Additionally, sample output images can be found in Fig. \ref{fig:results_hwy} and Fig. \ref{fig:results_phx}.  
Moreover, we perform ablation studies to examine the factors influencing the performance of our model, as well as comparing with the current state-of-the-art lane detector of \cite{sensetime}.

\begin{table*}[]
\centering
\begin{tabular}{|c|c|c||c|c|c|c|c|c|}
\hline
 Scenario & Camera & LiDAR  & DT L2 ($cm^2$) & DT L1 ($cm$) & AP & Pre. @25cm & Rec. @25cm & Top. Mean Dev. \\ \hhline{|=|=|=#=|=|=|=|=|=|} 
Highway & &  \checkmark    & 77.8  & 2.66 & 82.9\% & 95.0\% & 94.0\% & 0.337       \\ \hline
Highway & \checkmark &  & 110.4  &  3.35   & 78.8\% &  91.3\%  & 89.2\% &  0.446 \\ \hline
Highway & \checkmark  & \checkmark  & \textbf{70.5}  &  \textbf{2.63}   & \textbf{84.0\%} & \textbf{95.6\%} & \textbf{94.3\%} & \textbf{0.306}  \\ \hhline{|=|=|=#=|=|=|=|=|=|}

City & &  \checkmark    & 111.7  & \textbf{3.56} & 76.7\% & 88.3\% & \textbf{79.0\%} & 1.162       \\ \hline
City & \checkmark &  & 130.35  &  4.68   & 75.6\% &  86.3\%  & 72.4\% & 1.256  \\ \hline
City & \checkmark  & \checkmark  & \textbf{109.3}  &  \textbf{3.56}   & \textbf{78.1\%} & \textbf{89.3\%} & 76.9\% & \textbf{1.053}  \\ \hline

\end{tabular}
\caption{Comparisons of test set performance of lane detection results of using various sensors. }\label{sensor_comparison}
\end{table*}

\begin{table*}[]
\centering
\begin{tabular}{|c||c|c|c|c|c|c|}
\hline
                & \multicolumn{2}{c|}{AP}  & \multicolumn{2}{c|}{Pre @ 25cm} & \multicolumn{2}{c|}{Rec @ 25cm} \\ \hline
Lane Config     & SCNN   & Ours            & SCNN       & Ours               & SCNN       & Ours               \\ \hline
2 lanes         & 64.7\% & \textbf{85.5\%} & 75.2\%     & \textbf{96.9\%}    & 76.1\%     & \textbf{95.4\%}    \\ \hline
3 lanes         & 48.1\% & \textbf{78.6\%} & 54.3\%     & \textbf{90.2\%}    & 45.5\%     & \textbf{83.8\%}    \\ \hline
4 lanes         & 67.9\% & \textbf{84.2\%} & 78.4\%     & \textbf{96.7\%}    & 52.6\%     & \textbf{95.5\%}    \\ \hline
5 lanes         & 66.2\% & \textbf{82.8\%} & 76.3\%    & \textbf{95.3\%}    & 57.8\%     & \textbf{94.0\%}    \\ \hline
6 lanes or more & 66.4\% & \textbf{82.7\%} & 75.7\%     & \textbf{94.3\%}    & 46.8\%     & \textbf{93.1\%}    \\ \hline
SCNN compatible & 59.5\% & \textbf{84.7\%} & 69.1\%     & \textbf{96.3\%}    & 72.3\%     & \textbf{94.3\%}    \\ \hline
\end{tabular}
\caption{Comparison of test set performance of lane detection with the SCNN model and our LiDAR+Camera model in scenarios with different numbers of lanes. }\label{scnn}
\end{table*}

\begin{table*}[]
\centering
\begin{tabular}{|c|c||c|c|c|c|c|c|}
\hline
Scenario & Ground Height & DT L2 ($cm^2$) & DT L1 ($cm$) & AP & Pre. @25cm & Rec. @25cm & Top. Mean Dev. \\ \hhline{|=|=#=|=|=|=|=|=|}
Highway & Predicted  & \textbf{110.4}  &  \textbf{3.35}   & \textbf{78.8\%} &  \textbf{91.3\%}  & \textbf{89.2\%} & 0.446       \\ \hline
Highway & Ground Truth & 112.9  & 3.55 & 78.4\% & 91.0\% & 88.4\% & \textbf{0.391} \\ \hhline{|=|=#=|=|=|=|=|=|}

City & Predicted  & 130.4  &  4.68   & \textbf{75.6\%} &  \textbf{86.3\%}  & 72.4\% & 1.256  \\ \hline
City & Ground Truth & \textbf{124.8}  & \textbf{3.86} & 74.7\% & 85.7\% & \textbf{73.4\%} & \textbf{1.210} \\ \hline

\end{tabular}
\caption{Comparisons of lane detection results using only re-projected camera image onto ground truth ground height vs predicted ground height. }\label{height_comparison}
\end{table*}

\paragraph{Sensor Input} we  explore the performance impact of using information from the camera image, LiDAR, and the combination of both in Table \ref{sensor_comparison}. It is clear that the performance of the model that leverages both the LiDAR and camera information achieves the highest performance. This is especially true in the highway setting with faster vehicle motion. As a result, the distant regions still have relatively sparse LiDAR points despite aggregating multiple sweeps, while the much denser camera image is able to somewhat compensate for this. In comparison, the LiDAR-only model performs somewhat worse, while the camera-based model returns the lowest performance. 

\paragraph{Comparison with state of the art} we compare the lane detection results using the SCNN model \cite{sensetime} with ours, with the results in Table \ref{scnn}. \camready{In particular, we train the SCNN model using the authors' provided code on the camera images (as by their design) to produce lane detections in camera space. These detections are then re-projected into overhead view using the ground truth dense ground height and evaluated.} While  SCNN can only detect the ego lane and the neighboring left and right lanes, our model is able to detect all visible lanes within the 48 meter square. As such, the recall metric for SCNN is low in situations with more lanes. We showcase the performance difference across scenes with varying number of lanes. Because the SCNN detection receives only the image as input which cannot see the ground immediately next to the ego-vehicle, we ignore lanes within 15m of the ego-vehicle for both methods for a fair comparison. Finally, the last row of the table shows the methods evaluated over only scenes that contain an ego lane and at most one lane to the left and right, which is fully within the stated capabilities of SCNN. Our method signifiantly outperforms SCNN in all situations, further validating the effectiveness of our 3D reasoning. 

\paragraph{Evaluation over Distance} because of the foreshortening effect, the density of measurement points falling onto the ground falls off rapidly with the distance from the vehicle. Since LiDAR sweeps have much lower resolution than a camera image, this effect is more pronounced in the former case. To further analyze the benefit of using both camera and LiDAR data compared with using only the latter, we plot the relative average precision of the two models versus distance from the vehicle, as shown in Fig. \ref{fig:distance}. It is clear that in both scenarios, the performance of the former model exceeds that of the latter. This performance gap is especially large at the longest distance bin, where the number of LiDAR points on the ground is very sparse even with sweep aggregation. This suggests that intelligently combining the information from both sensors will provide the best lane detections. 

\paragraph{Ground height prediction} our model is able to predict the ground height onto which the camera images are projected. In Table \ref{height_comparison} we explore the performance gains possible if the accuracy of ground height estimation is improved by substituting the predictions with ground truth.  For this experiment, we only provide the LiDAR BEV image to the ground height estimator so that the model can only perform lane detection using the re-projected camera image. We see that the performance is very similar when using estimated and perfect ground. 

\paragraph{Qualitative Analysis} in Fig. \ref{fig:results_hwy} and Fig. \ref{fig:results_phx}, we see a number of examples where the lane detections are very accurate. The high level of performance is achieved in a variety of scenarios including with complex lane geometries (large number of lanes or merges and exits), as well as occlusions by other vehicles. Moreover, the alignment of the lanes in both the camera view and in BEV are seen to be of high quality. The performance of the lane detector in adverse scenarios in Fig. \ref{fig:results_hwy} such as darkness, rain, and fog suggests that the model is able to perform in a variety of situations. Finally, the smoothness of the predicted lane boundaries is noteworthy, and highly beneficial to the consumers of the detection results. 


\paragraph{Failure modes} despite the generally very accurate lane detections, there are a few failure modes of our model. These can be seen in Fig. \ref{fig:results_failure}. The first row shows a case where the heavy rain causes significantly reduced visibility, especially at a distance. In the second row, large vehicles block the view of the ground for both sensors, showing the negative impact of heavy occlusions. Finally, the third image shows a case where the lane boundaries do not have actual paint in reality. In these scenarios, more advanced models are required to infer the virtual lane boundaries.



\section{Conclusion}


 In this paper we have argued that accurate image estimates do not translate to precise 3D lane boundaries, which are  the input required by modern motion planning algorithms. To address this issue, we have proposed a novel  deep neural network  that  takes advantage of both LiDAR and camera sensors and  produces very accurate estimates directly in 3D space. We have demonstrated the performance of our approach on two challenging real world datasets containing both  highway and city scenes, showing its superiority when compared to the state of the art. In the future we plan to reason about lane attributes such as lane types and other road elements such as crosswalks.



{\small
\bibliographystyle{IEEEtran}
\bibliography{ref}
}

\end{document}